\documentclass[10pt,twocolumn,letterpaper]{article}

\usepackage{cvpr}              

\usepackage{graphicx}
\usepackage{amsmath}
\usepackage{amssymb}
\usepackage{booktabs}
\usepackage{svg}

\usepackage[pagebackref,breaklinks,colorlinks]{hyperref}

\usepackage[capitalize]{cleveref}
\crefname{section}{Sec.}{Secs.}
\Crefname{section}{Section}{Sections}
\Crefname{table}{Table}{Tables}
\crefname{table}{Tab.}{Tabs.}

\def\cvprPaperID{4840} 
\def\confName{CVPR}
\def\confYear{2023}

\begin{document}

\title{VecFontSDF: Learning to Reconstruct and Synthesize High-quality Vector Fonts via Signed Distance Functions}

\author{Zeqing Xia\footnotemark[1]~, Bojun Xiong\footnotemark[1]~, Zhouhui Lian\footnotemark[2]\\
Wangxuan Institute of Computer Technology, Peking University, China\\
}
\maketitle

\renewcommand{\thefootnote}{\fnsymbol{footnote}}
\footnotetext[1]{Denotes equal contribution.}
\footnotetext[2]{Corresponding author. E-mail: lianzhouhui@pku.edu.cn\\This work was supported by National Language Committee of China (Grant No.: ZDI135-130), Center For Chinese Font Design and Research, and Key Laboratory of Science, Technology and Standard in Press Industry (Key Laboratory of Intelligent Press Media Technology).}

\begin{abstract}
Font design is of vital importance in the digital content design and modern printing industry. Developing algorithms capable of automatically synthesizing vector fonts can significantly facilitate the font design process. However, existing methods mainly concentrate on raster image generation, and only a few approaches can directly synthesize vector fonts. This paper proposes an end-to-end trainable method, VecFontSDF, to reconstruct and synthesize high-quality vector fonts using signed distance functions (SDFs). Specifically, based on the proposed SDF-based implicit shape representation, VecFontSDF learns to model each glyph as shape primitives enclosed by several parabolic curves, which can be precisely converted to quadratic Bézier curves that are widely used in vector font products. In this manner, most image generation methods can be easily extended to synthesize vector fonts. Qualitative and quantitative experiments conducted on a publicly-available dataset demonstrate that our method obtains high-quality results on several tasks, including vector font reconstruction, interpolation, and few-shot vector font synthesis, markedly outperforming the state of the art. Our code and trained models are available at \url{https://xiazeqing.github.io/VecFontSDF}.
\end{abstract}

\begin{figure}[t!]
  \includegraphics[width=.95\columnwidth]{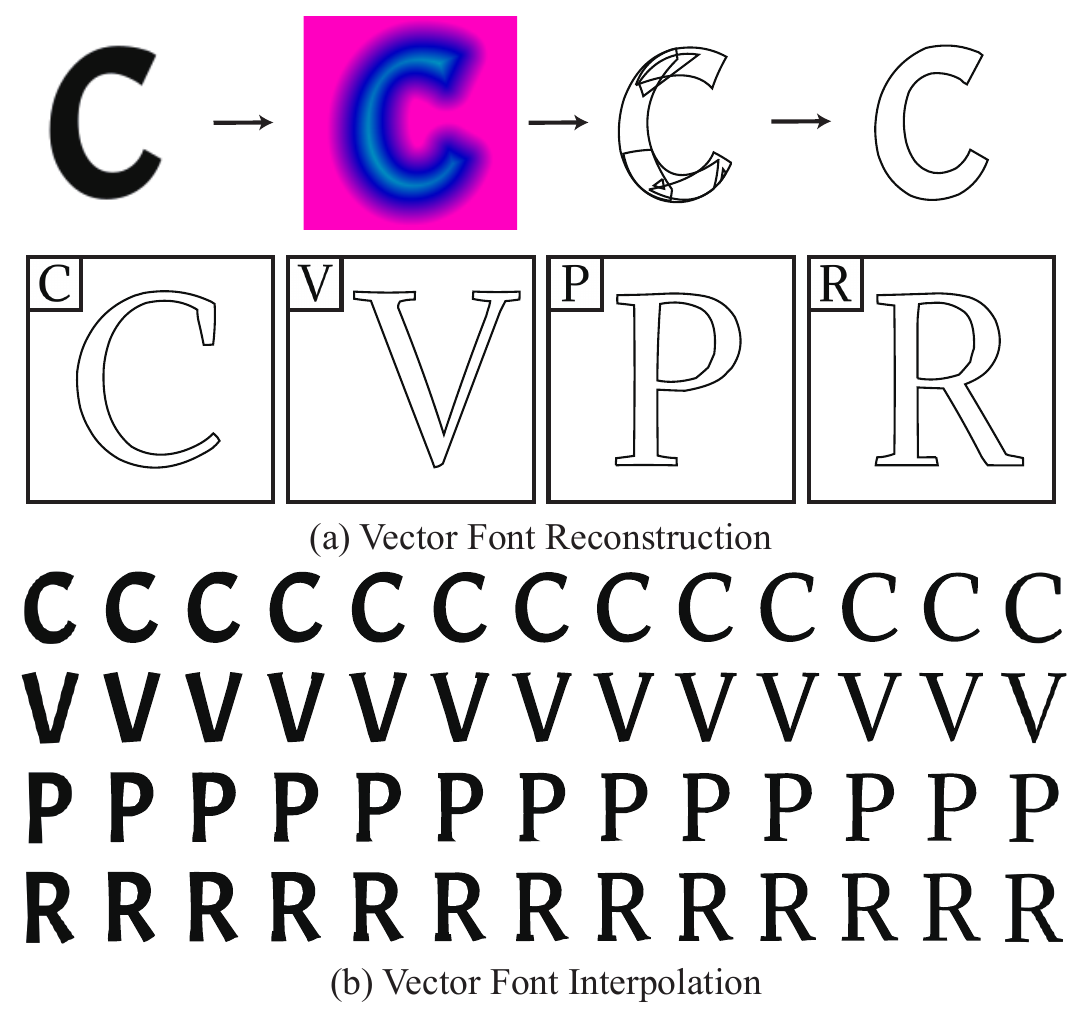}
  \caption{Examples of results obtained by our method in the tasks of vector font reconstruction (a) and vector font interpolation (b).}
  \label{fig:teaser}
\end{figure}

\section{Introduction}
\label{sec:intro}

Traditional vector font designing process relies heavily on the expertise and effort from professional designers, setting a high barrier for common users. With the rapid development of deep generative models in the last few years, a large amount of effective and powerful methods \cite{azadi2018multi, wang2020attribute2font, gao2019artistic} have been proposed to synthesize visually-pleasing glyph images. In the meantime, how to automatically reconstruct and generate high-quality vector fonts is still considered as a challenging task in the communities of Computer Vision and Computer Graphics. Recently, several methods based on sequential generative models \cite{carlier2020deepsvg, DavidHa2017ANR, lopes2019learned,tang2019fontrnn, wang2021deepvecfont} have been reported that treat a vector glyph as a draw-commands sequence and use Recurrent Neural Networks (RNNs) or Transformer \cite{vaswani2017attention} to encode and decode the sequence. However, this explicit representation of vector graphics is extremely difficult for the learning and comprehension of deep neural networks, mainly due to the long-range dependence and the ambiguity in how to draw the outlines of glyphs. More recently, DeepVecFont \cite{wang2021deepvecfont} was proposed to use dual-modality learning to alleviate the problem, showing state-of-the-art performance on this task. Its key idea is to use a CNN encoder and an RNN encoder to extract features from both image and sequence modalities. Despite using richer information of dual-modality data, it still needs repetitively random samplings of synthesis results to find the optimal one and then uses Diffvg \cite{li2020differentiable} to refine the vector glyphs under the guidance of generated images in the inference stage.

Another possible solution to model the vector graphic is to use implicit functions which have been widely used to represent 3D shapes in recent years. For instance,  DeepSDF~\cite{DBLP:conf/cvpr/ParkFSNL19} adopts a neural network to predict the values of signed distance functions for surfaces, but it fails to convert those SDF values to the explicit representation. BSP-Net~\cite{chen2020bspnet} uses hyperplanes to split the space in 2D shapes and 3D meshes, but it generates unsmooth and disordered results when handling glyphs consisting of numerous curves. 

Inspired by the above-mentioned existing methods, we propose an end-to-end trainable model, VecFontSDF, to reconstruct and synthesize high-quality vector fonts using signed distance functions (SDFs). The main idea is to treat a glyph as several primitives enclosed by parabolic curves which can be translated to quadratic Bézier curves that are widely used in common vector formats like SVG and TTF. Specifically, we use the feature extracted from the input glyph image by convolutional neural networks and decode it to the parameters of parabolic curves. Then, we calculate the value of signed distance function for the nearest curve of every sampling point and leverage these true SDF values as well as the target glyph image to train the model.

The work most related to ours is \cite{9797843}, which also aims to provide an implicit shape representation for glyphs, but possesses a serious limitation mentioned below. For convenience, we name the representation proposed in~\cite{9797843} ``IGSR", standing for ``Implicit Glyph Shape Representation". The quadratic curves used in IGSR are not strictly parabolic curves, which cannot be translated to quadratic Bézier curves. As a consequence, their model is capable of synthesizing high-resolution glyph images but not vector glyphs. Furthermore, it only uses raster images for supervision which inevitably leads to inaccurate reconstruction results. On the contrary, our proposed VecFontSDF learns to reconstruct and synthesize high-quality vector fonts that consist of quadratic Bézier curves by training on the corresponding vector data in an end-to-end manner. Major contributions of our paper are threefold:
\begin{itemize}
\item We design a new implicit shape representation to precisely reconstruct high-quality vector glyphs, which can be directly converted into commonly-used vector font formats (e.g., SVG and TTF).
\item We use the true SDF values as a strong supervision instead of only raster images to produce much more precise reconstruction and synthesis results compared to previous SDF-based methods. 
\item The proposed VecFontSDF can be flexibly integrated with other generative methods such as latent space interpolation and style transfer. Extensive experiments have been conducted on these tasks to verify the superiority of our method over other existing approaches, indicating its effectiveness and broad applications.
\end{itemize}

\section{Related Work}

\subsection{Vector Font Generation}
In early years, researchers tried to generate glyph images and utilized traditional vectorization methods~\cite{kolesnikov2007polygonal, pan2014skeleton} to obtain vector fonts. With the development of sequential models such as long short-term memory RNNs \cite{hochreiter1997long} and Transformer~\cite{vaswani2017attention}, a number of methods have been developed that treat vector graphics as drawing-command sequences for modeling.
Lopes et al. \cite{lopes2019learned} proposed SVG-VAE that provides a scale-invariant representation for imagery and then uses an RNN decoder to generate vector glyphs. Carlier et al.\cite{carlier2020deepsvg} proposed a novel hierarchical generative network called DeepSVG, which effectively disentangles high-level shapes from the low-level commands that encode the shape itself and directly predicts a set of shapes in a non-autoregressive fashion. Reddy et al.\cite{reddy2021im2vec} proposed Im2Vec that can generate complex vector graphics with varying topologies, and only requires indirect supervision from raster images. Mo et al.\cite{mo2021general} introduced a general framework to produce line drawings from a wide variety of images by using a dynamic window. At the same time, Wang et al. \cite{wang2021deepvecfont} adopted dual-modality learning to synthesize visually pleasing vector glyphs. However, all the methods mentioned above are based on the explicit representation of vector fonts, which requires the neural networks to learn long-range dependence and generate the drawing commands step by step. On the contrary, our method regards the vector glyph as a whole and is trained by the deterministic values of signed distance functions (SDFs).

\begin{figure*}[!t]
  \centering
  \includegraphics[width=\textwidth]{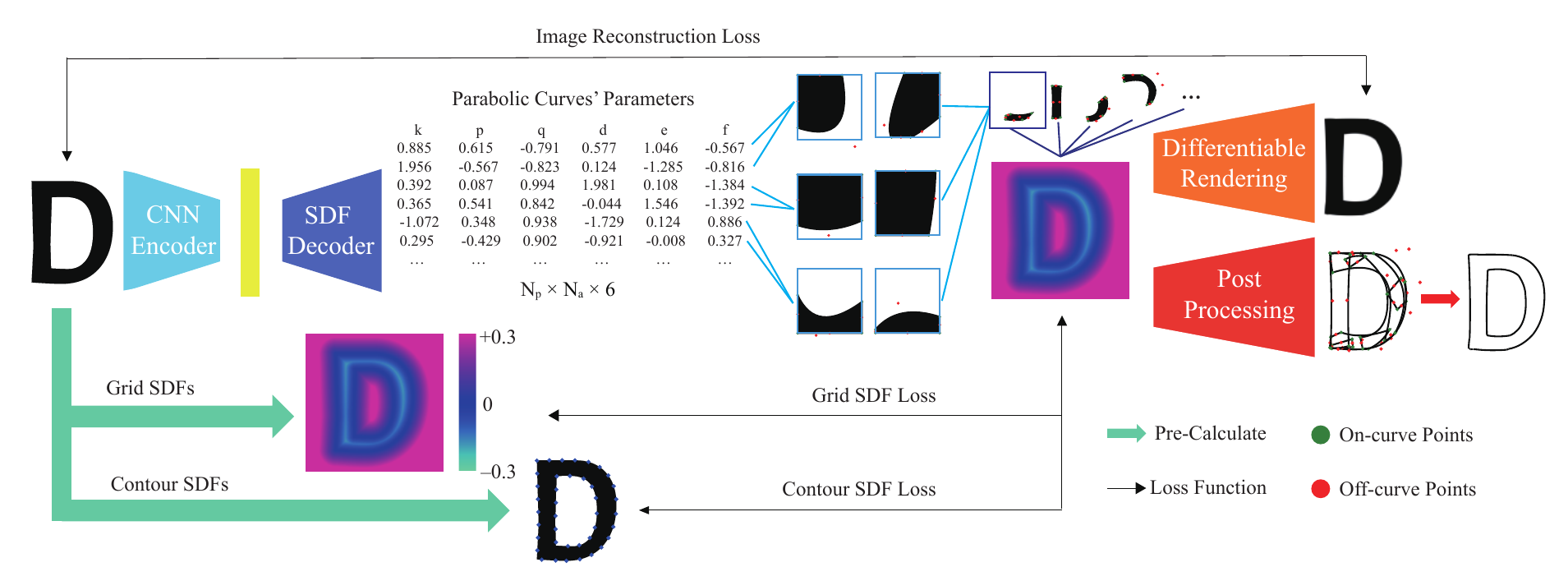}
  \caption{An overview of our vector font reconstruction framework.}
  \label{fig:overview}
\end{figure*}

\subsection{Implicit Shape Representation}

In the last two decades, lots of distance functions-based algorithms have been developed for implicit shape representation. As a pioneering work, Gibson et al.~\cite{DBLP:conf/siggraph/FriskenPRJ00} proposed an adaptively sampled distance field using octree for both 2D glyph representation and 3D rendering. Green~\cite{DBLP:conf/siggraph/Green07a} used distance functions for glyph rasterization. With the rapid development of machine learning techniques, signed distance functions and other implicit functions-based methods have been widely used in 2D and 3D reconstruction tasks. Park et al.~\cite{DBLP:conf/cvpr/ParkFSNL19} proposed DeepSDF which uses an auto-decoder to predict the values of signed distance functions for 3D reconstruction. Chen et al.~\cite{DBLP:conf/cvpr/ChenZ19} presented an implicit function using fully-connected layers that output signed distances with the input of image features and sampling points. However, the implicit representations they used can not be directly converted to traditional explicit descriptions.

Aiming at 3D shape decomposition, Deng et al.~\cite{DBLP:conf/cvpr/DengGYBHT20} proposed CVXNet and Chen et al.~\cite{chen2020bspnet} proposed BSP-Net, which both employ hyperplanes to split the space to get convex hulls. However, such methods only work well on simple convex objects, performing poorly on shapes with concave curves. Liu et al.~\cite{9797843} replaced the hyperplanes with quadratic curves. It does successfully reconstruct glyph images with an arbitrary resolution, but cannot convert its output to conventional Bézier curves. On the contrary, our method is able to reconstruct and generate visually-pleasing vector fonts using a new explicit shape representation based on our proposed pseudo distance functions.

\section{Method Description}
\label{sec:method_overview}
In order to directly reconstruct and synthesize vector fonts from input glyph images via neural networks, we need a learnable shape representation that precisely preserves geometry information. Previous methods that fully rely on sequential generative models such as SVG-VAE \cite{lopes2019learned} and DeepVecFont \cite{wang2021deepvecfont} fail to perfectly handle this task mainly due to the ambiguity of control point selection. To address this problem, we propose a vector graphic reconstructor based on an interpretable distance function with differentiable rendering. Unlike most SDF-based methods such as DeepSDF \cite{smirnov2020deep} which directly predicts the values of signed distance functions via neural networks, our model outputs the learnable parameters of curves that can be applied to handle complex glyph outlines containing concave curves.

Fig.~\ref{fig:overview} shows the pipeline of our VecFontSDF. For data preparation, we need to pre-calculate the values of signed distance functions (SDFs) for each input glyph using the method described in Sec.~\ref{sec:sdf}. We have two types of pre-calculated SDFs: grid SDFs and contour SDFs. Grid SDFs mean that the sampling points used to calculate SDFs are located at all grid positions (i.e., pixels in a raster image). Contour SDFs mean that the sampling points are uniformly distributed near the contours of the input glyph. Then, a CNN encoder is trained to extract the features from input raster images and an SDF decoder is followed to predict the parameters for every parabolic curve. We design a new pseudo distance function to calculate the distance from every sampling point to the parabolic curve based on the predicted parameters. The values of pseudo distance functions are supervised by real SDFs to achieve more precise reconstruction results. Furthermore, with a simple but effective differentiable renderer, SDFs can be converted into raster images which are compared with input glyph images. Finally, the above parameters of parabolic curves are converted into quadratic Bézier curves after the post-processing step. More details of our VecFontSDF are described in the following subsections.

\subsection{Signed Distance Functions}
\label{sec:sdf}

We consider a glyph $C=\left\{c_1,c_2,...,c_{N_c}\right\}$ to be a set of $N_c$ contours $c_i$, and each $c_i$ is defined as a sequence of Bézier curves $c_i=\{B_1,B_2,...,B_{N_B}\}$, where $N_B$ indicates the amount of Bézier curves in the sequence $c_i$. To generate practical vector font libraries (e.g., TTF fonts), our model adopts the quadratic Bézier curve $B$ which is defined as:
\begin{equation}
    B:{P(t)}=(1-t)^2P_0+2t(1-t)P_1+t^2P_2, 0\leq t\leq 1,
    \label{equ:basic_bezier}
\end{equation}
where $P_0$ and $P_2$ denote the first and last control points (on-curve points), respectively, and $P_1$ is the intermediate (off-curve) control point (see Fig. ~\ref{fig:overview}).

To calculate the signed distance function of a given point $P(x,y)$ towards the glyph outline, we first need to determine its sign, and then find its nearest curve $\hat{B}$ and the corresponding parameter $\hat{t}$:
\begin{equation}
\begin{split}
    \hat{t}&={\rm argmin}_t{{\left|\left|P(t)-P\right|\right|}_2^2} \\
    {\left|\left|P(t)-P\right|\right|}_2^2&=(x(t)-x)^2+(y(t)-y)^2 \triangleq f^4(t).
\end{split}
\end{equation}
Since the distance function is a quartic function, we get its minimum point by letting its derivative equal to zero. After obtaining $\hat{t}$, we can calculate the distance value:
\begin{equation}
    {\rm dist}^2(\hat{t})={\left|\left|P(\hat{t})-P\right|\right|}_2^2=(x(\hat{t})-x)^2+(y(\hat{t})-y)^2,
\end{equation}
and its sign:
\begin{equation}
        {\rm sgn}(\hat{t})=\frac{\overrightarrow{P(\hat{t})P}\times\overrightarrow{v(\hat{t})}}{\left|\overrightarrow{P(\hat{t})P}\times\overrightarrow{v(\hat{t})}\right|},
\end{equation}
where $v(t)$ denotes the derivative of $P(t)$:
\begin{equation}
    v(\hat{t})=-2(1-\hat{t})P_0+2(1-2\hat{t})P_1+2\hat{t}P_2.
\end{equation}

\begin{figure}[t!]
  \centering
  \includegraphics[width=\columnwidth]{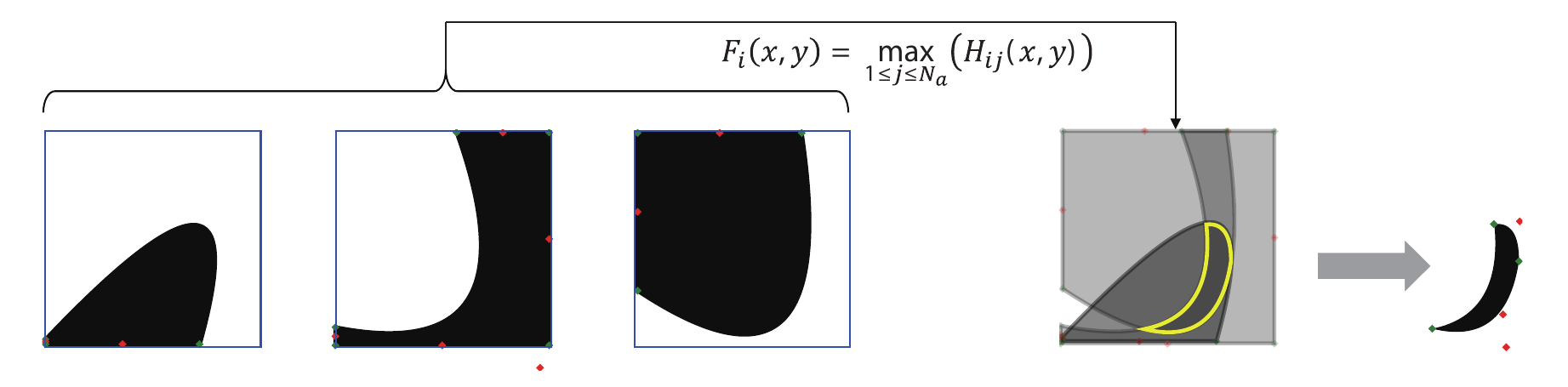}
  \caption{An illustration of calculating the intersection of $N_a$ areas via maximum operator.}
  \label{fig:method_pesudo_intersection}
\end{figure}

Finally, the signed distance is computed by:
\begin{equation}
        D(x,y)={\rm dist}(\hat{t})\times{\rm sgn}(\hat{t}).
    \label{equ:real_sdf}
\end{equation}

\begin{figure*}[!t]
  \centering
  \includegraphics[width=\textwidth]{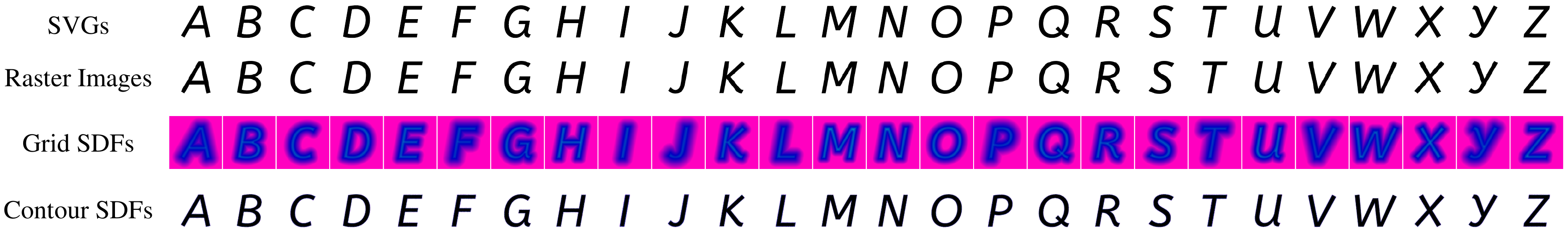}
  \caption{Examples of vector glyphs, glyph images, grid SDFs and contour SDFs. For better visualization of grid SDFs, we map the positive value of distance functions to the red channel of RGB images and map the negative value to the green channel. To visualize the contour SDFs, we add all the contour sampling points into the original SVG files. Please zoom in for better inspection.}
  \label{fig:dataset}
\end{figure*}

To fully exploit the glyph's vector information, we repeat the above calculation on all grid positions and uniformly sample points near glyph contours to obtain grid SDFs and contour SDFs, respectively. An illustration of the proposed grid SDFs and contour SDFs can be found in Fig.~\ref{fig:dataset}.

\subsection{Pseudo Distance Functions}
Due to the complex gradient back propagation process, directly calculating the real signed distances from sampling points to parabolic curves via Eq.~\ref{equ:real_sdf} is infeasible when training our neural network. To solve this problem, we propose a pseudo distance function inspired by the algebraic distance \cite{DBLP:journals/tog/GuennebaudG07} to simulate the original signed distance function. As mentioned above, our model aims to generate a set of parabolic curves which are defined by:
\begin{equation}
    k(px+qy)^2+dx+ey+f=0.
    \label{equ:pseudo_dis_equ}
\end{equation}
Thus, we can define the pseudo distance function of a sampling point to a parabolic curve as:
\begin{equation}
    H(x,y)=k(px+qy)^2+dx+ey+f.
    \label{equ:pseudo_dis_func}
\end{equation}

Similar to the original signed distance function, our pseudo distance function also regards the point with a positive distance value as outside and vice versa. The area inside a parabolic curve is defined as $H(x,y) < 0$. To reconstruct the geometry of an input glyph image (see Fig.~\ref{fig:method_pesudo_intersection}), we compute the intersection of $N_a$ areas to get a shape primitive:
\begin{equation}
    F_i(x,y)=\max_{1\leq j\leq N_a}(H_{ij}(x,y)) \quad
    \left\{
    \begin{array}{rcl}
    >0 & \rm{outside} \\
    \leq0 & \rm{inside}
    \end{array},
    \right.
\end{equation}
where $F_i(x,y) < 0$ denotes the i-th primitive, and $N_p$ denotes the number of primitives used to reconstruct the glyph image. An illustration of the above maximum operator is shown in Fig.~\ref{fig:method_pesudo_intersection}.

\begin{figure}[t!]
  \centering
  \includegraphics[width=\columnwidth]{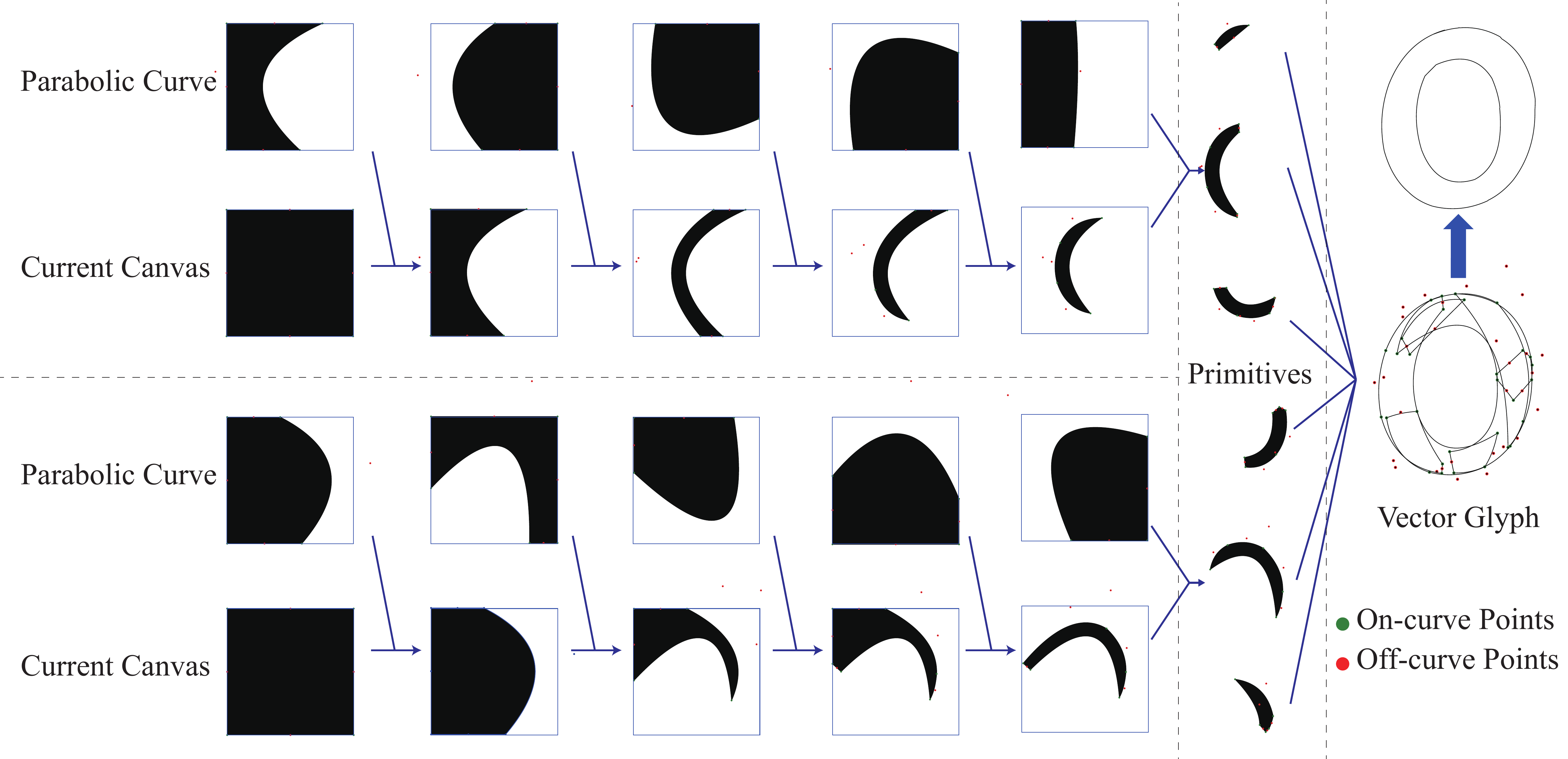}
  \caption{A demonstration of our post processing step.}
  \label{fig:method_postproc}
\end{figure}

Finally, we get the proposed pseudo distance field by:
\begin{align}
    G&=\min_{1\leq i\leq N_p}(F_i)
    \quad
    \left\{
    \begin{array}{rcl}
    >0 & \rm{outside} \\
    \leq0 & \rm{inside}
    \end{array},
    \right.
    \label{equ:pseudo_sdf}
\end{align}
where $G_i < 0$ denotes the combination of $N_p$ primitives. Thus, our model outputs $N_p \times N_a$ parabolic curves where each curve has 6 parameters $\{k,p,q,d,e,f\}$ as defined in Eq.~\ref{equ:pseudo_dis_equ}. 
Our representation is able to depict the concave regions of a glyph due to the utilization of the parameter $k$ whose sign determines whether a region is inside or outside. 

\subsection{End-to-End Vector Reconstruction}
\label{sec:train}

After obtaining the final pseudo distance field $G$, we need to render it to the raster image in an differentiable manner. Inspired by~\cite{NEURIPS2021_6948bd44}, we rasterize $G\in(-\infty,\infty)$ to the glyph image $\hat{I}\in[0,1]$ by computing:
\begin{small}
\begin{align}
    \hat{I}(g_{x,y})=\left\{
    \begin{array}{ll}
         1 & \gamma < g_{x,y} \\
         \frac{1}{2}-\frac{1}{4}\left(\left(\frac{g_{x,y}}{\gamma}\right)^3-3\left(\frac{g_{x,y}}{\gamma}\right)\right)& -\gamma \leq g_{x,y} \leq \gamma \\
         0 & g_{x,y} < -\gamma
    \end{array},
    \right.
    \label{equ:render}
\end{align}
\end{small}
where $\gamma$ represents the learnable range of SDF values. Only the position whose pseudo distance value belongs to $[-\gamma, \gamma]$ can back-propagate its gradient to update the network.

Our model is trained by minimizing the loss function that consists of the image reconstruction loss $L_{image}$, the grid SDF loss $L_{grid}$, the contour SDF loss $L_{contour}$, and the regularization loss $L_{regular}$. The image reconstruction loss is defined as the mean square error between the reconstructed image $\hat{I}$ and the input image $I$:
\begin{equation}
    L_{image}=\left|\left|\hat{I}-I\right|\right|_2^2.
\end{equation}

To further exploit the vector information, we use the pre-calculated grid SDFs and contour SDFs to provide a much more precise supervision. However, due to the inconsistency of calculation between the real distance field $D$ from Eq.~\ref{equ:real_sdf} and the pseudo distance field $G$ from Eq.~\ref{equ:pseudo_sdf}, we cannot directly calculate their numerical difference. But since they share the same zero level-set and monotony with the movement of sampling points, we only calculate the loss when they have different signs, which means the predicted curve and the ground-truth curve are on the opposite side of a sampling point. The contour SDF loss is defined as:
\begin{equation}
    L_{contour}=\frac{1}{M_{c}}\sum_{x,y\in S_{contour}}{{\rm ReLU}\left(-G(x,y)\times D(x,y)\right),}
\end{equation}
where $S_{contour}$ denotes the set containing sampling points uniformly distributed near the glyph contours as described in Sec.~\ref{sec:sdf}, and $M_{c} = |S_{contour}|$. 

Similarly, the grid SDF loss $L_{grid}$ can be computed by:
\begin{equation}
    L_{grid}=\frac{1}{M_{g}}\sum_{x,y\in S_{grid}}{{\rm ReLU}\left(-G(x,y)\times D(x,y)\right),}
\end{equation}
where $S_{grid}$ denotes the set consisting of $M_{g}=W\times H$ points sampled from all pixels of input image with the width $W$ and height $H$.

To further improve the quality of reconstructed images, we also introduce the regularization loss $L_{regular}$ to our model. Specifically, we first need to normalize the parameters $p$ and $q$ used in Eq.~\ref{equ:pseudo_dis_func} by forcing $p^2+q^2=1$. Then, we restrict the minimum value of $\hat{k}^2$ outputted by our model greater than a pre-defined $k^2$ to obtain a clear image. The regularization loss $L_{regular}$ is defined as:
\begin{equation}
\begin{split}
    L_{regular}&=\frac{1}{N_p \times N_a}\left(\lambda_{k^2}\sum{{\rm ReLU}\left(k^2-\hat{k}^2\right)}\right. \\
    &\left.+\sum{\left(p^2+q^2-1\right)^2}\right).
\end{split}
\end{equation}

Finally, the complete loss function of our model is defined as the weighted sum of all above losses:
\begin{equation}
\begin{split}
    L_{total}&=\lambda_{image}L_{image}+\lambda_{grid}L_{grid} \\
    &+\lambda_{contour}L_{contour}+\lambda_{regular}L_{regular}.
\end{split}
\label{equ:whole-loss}
\end{equation}

\subsection{Post Processing}
In order to synthesize vector fonts for practical uses, we need to convert the parameters of parabolic curves generated by our SDF decoder to quadratic Bézier curves. Fig. ~\ref{fig:method_postproc} illustrates the post-processing step of our method. For each primitive, our initial canvas starts from a square and each curve splits the space into an inside region and an outside region. Then, we calculate the intersection of the inside region of newly-added curve and the current canvas to update the canvas recursively to get the final primitive. Finally, we assemble all the primitives together to form the output vector glyph.  What's more, we further merge the outlines of all primitives to get a complete outline representation of the glyph. More details regarding how to calculate the control points of quadratic Bézier curves in those primitives and merge their outlines are shown in  supplementary materials.

\section{Experiments}
\label{sec:experiments}


We conduct both qualitative and quantitative experiments on the dataset released by~\cite{o2014exploratory}, which consists of 1,116 fonts. We use the first 1,000 fonts for training and the rest 116 for testing. We use the method described in Sec.~\ref{sec:method_overview} to convert the SVG file of each vector glyph to grid SDFs and contour SDFs. Meanwhile, we use the traditional rasterization method to render every vector glyph to the corresponding glyph image with the resolution of 128 $\times$ 128. Examples of vector glyphs, glyph images, grid SDFs and contour SDFs are shown in Fig.~\ref{fig:dataset}.

In our experiments, all the input and output images are in the resolution of $128 \times 128$. We calculate the grid SDFs on the center position of every pixel, i.e., $(0.5,0.5),(0.5,1.5),(0.5,2.5),...,(127.5,127.5)$ under our image resolution setting (i.e., $M_{g}=128 \times 128$). The total number of contour SDFs sampling points $M_{c}$ are set to 4,000. Locations of all the sampling points are normalized to [-1,1] during training for better stability. After normalization, the threshold $\gamma$ in Eq.~\ref{equ:render} is set to 0.02. The encoder to extract image feature used in our VecFontSDF is ResNet-18~\cite{he2016deep} with Leaky     Relu~\cite{maas2013rectifier} and Batch Normalization~\cite{ioffe2015batch}. For the implicit SDF decoder, we set $N_p = 16$ and $N_a = 6$ for each primitive. Weights in the loss function are selected as $\lambda_{image}=1$, $\lambda_{grid} = 100$, $\lambda_{contour}=1000$, $\lambda_{regular}=1$ and $\lambda_{k^2} = 0.1$. We utilize Adam~\cite{kingma2014adam} with $lr = 1e-4$ and $betas = (0.9, 0.999)$ as the optimizer. We train our vector glyph reconstruction model with the batch size 64 for 100,000 iterations.

\begin{figure}[t!]
  \centering
  \includegraphics[width=\columnwidth]{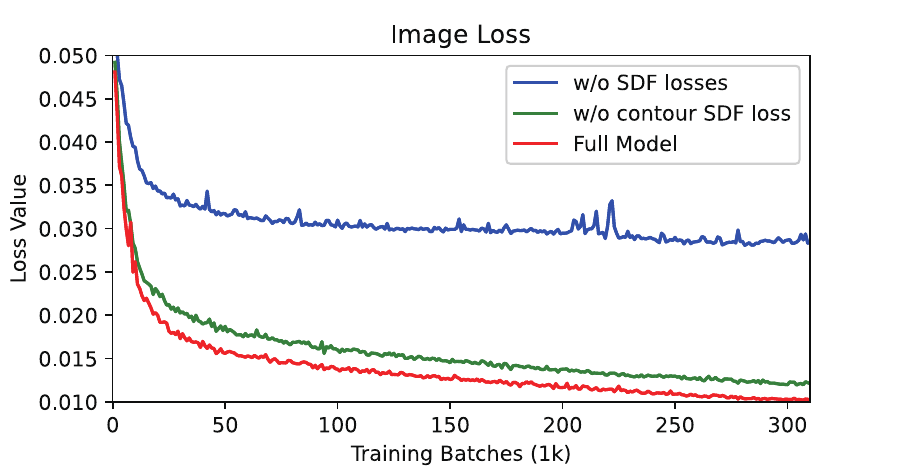}
  \caption{The loss curves of our model evaluated on the test set when training with different losses.}
  \label{fig:quantitative}
\end{figure}

\begin{table}[t!]
    \centering
    \caption{Quantitative results of our VecFontSDF using different losses evaluated on the test set for vector font reconstruction.}

    \label{tab:ablation}
    \centering%
    \resizebox{\columnwidth}{!}{
    \begin{tabular}{lccccc}
      \toprule
      Method & $\rm L_1$ distance $\downarrow$ & IoU$\uparrow$ & PSNR$\uparrow$ & LPIPS$\downarrow$ &  SSIM$\uparrow$ \\
      \midrule
      w/o SDF losses & 0.0445 & 0.9569 & 17.7928 & 0.2327 & 0.8620 \\
      w/o contour SDF loss & 0.0110 & 0.9876 & 25.9785 & 0.0384 & 0.9587 \\
      Full Model & \textbf{0.0090} & \textbf{0.9901} & \textbf{27.8502}  & \textbf{0.0303} & \textbf{0.9669} \\
      \bottomrule
    \end{tabular}}
\end{table}

\subsection{Ablation Study}
For the purpose of analyzing the impacts of different losses, we conduct a series of ablation studies by removing the corresponding loss terms used in our model on the vector reconstruction task.
The loss curves of $L_{image}$ on the test set shown in Fig.~\ref{fig:quantitative} demonstrate the effects of our proposed SDF losses. From the image loss curves, we witness a notable improvement brought by introducing the gird SDF loss. It's obvious that the grid SDF loss provides a much stronger supervision of the values of the output SDFs on every grid position. The raster images only contain the pixel value on every grid position where most of them are zeros or ones. As a consequence, using only the raster images to guide the training of networks losses a great deal of information. Moreover, introducing the contour SDF loss further improves the model's ability in the inference stage.

For further evaluation, we also apply several commonly-used metrics for image synthesis: $L_1$ distance, Intersection over Union (IoU), Peak Signal to Noise Ratio (PSNR), Learned Perceptual Image Patch Similarity (LPIPS)~\cite{zhang2018perceptual} and structural similarity (SSIM) ~\cite{1284395} distance between the reconstructed images rendered by output SDFs and input glyph images. Table~\ref{tab:ablation} shows that using all the losses achieves the best performance on the task of vector font reconstruction.

\begin{figure}[t!]
  \centering
  \includegraphics[width=\columnwidth]{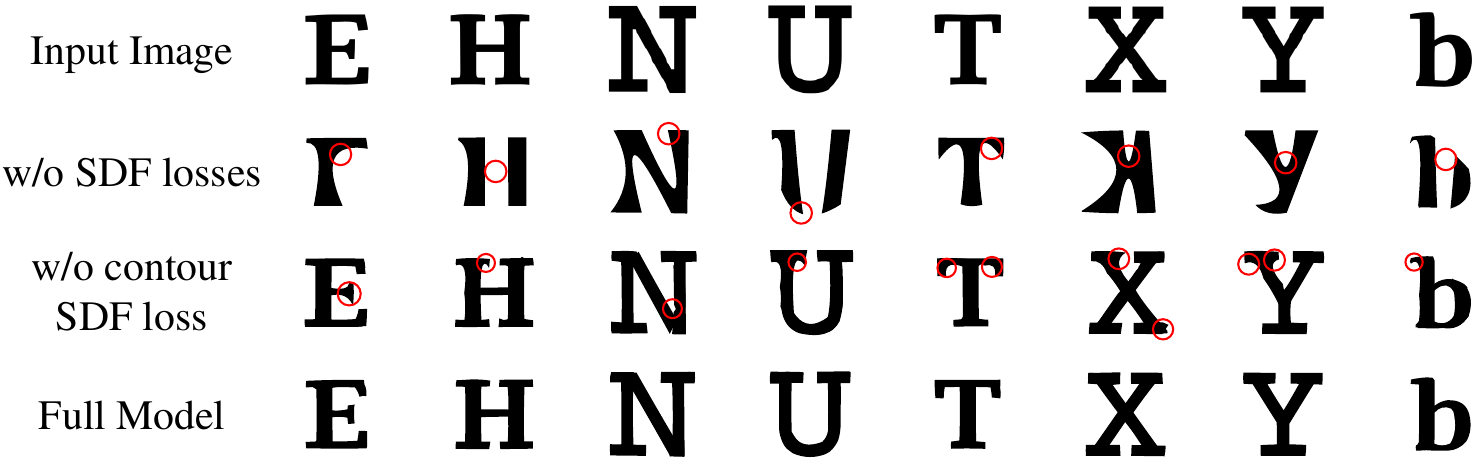}
  \caption{Qualitative results of our model using different losses. Red circles highlight the shortcomings of VecFontSDF variants.}
  \label{fig:qualitative}
\end{figure}

Fig.~\ref{fig:qualitative} provides some qualitative results to verify the effectiveness of each proposed loss. Simply using the image loss, our model even fails to synthesize the correct glyph shape since the raster image only provides the sign of distance functions on every grid point. Using the grid SDF loss results in a far more accurate supervision in leading the model to capture the geometry information of input images. But the model is still incapable of handling the details of glyph contours, e.g., the areas in red circles in Fig.~\ref{fig:qualitative}. The contour SDF loss successfully solves this problem and helps our model sufficiently learn these details of glyph contours, such as corners and serifs.

\begin{figure*}[t!]
  \centering
  \includegraphics[width=\textwidth]{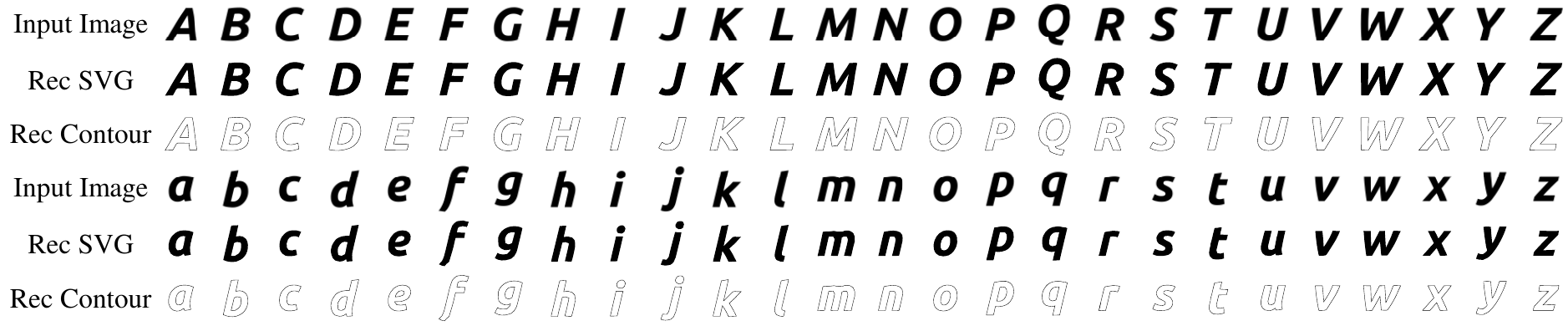}

  \caption{Examples of our vector font reconstruction results on the test set. ``Rec SVG" denotes the filled shapes of the reconstructed vector glyphs and ``Rec Contour" denotes the contours of corresponding vector glyphs. Please zoom in for better inspection.}
  \label{fig:rec}
\end{figure*}

\begin{figure}[t!]
  \centering
  \includegraphics[width=.9\columnwidth]{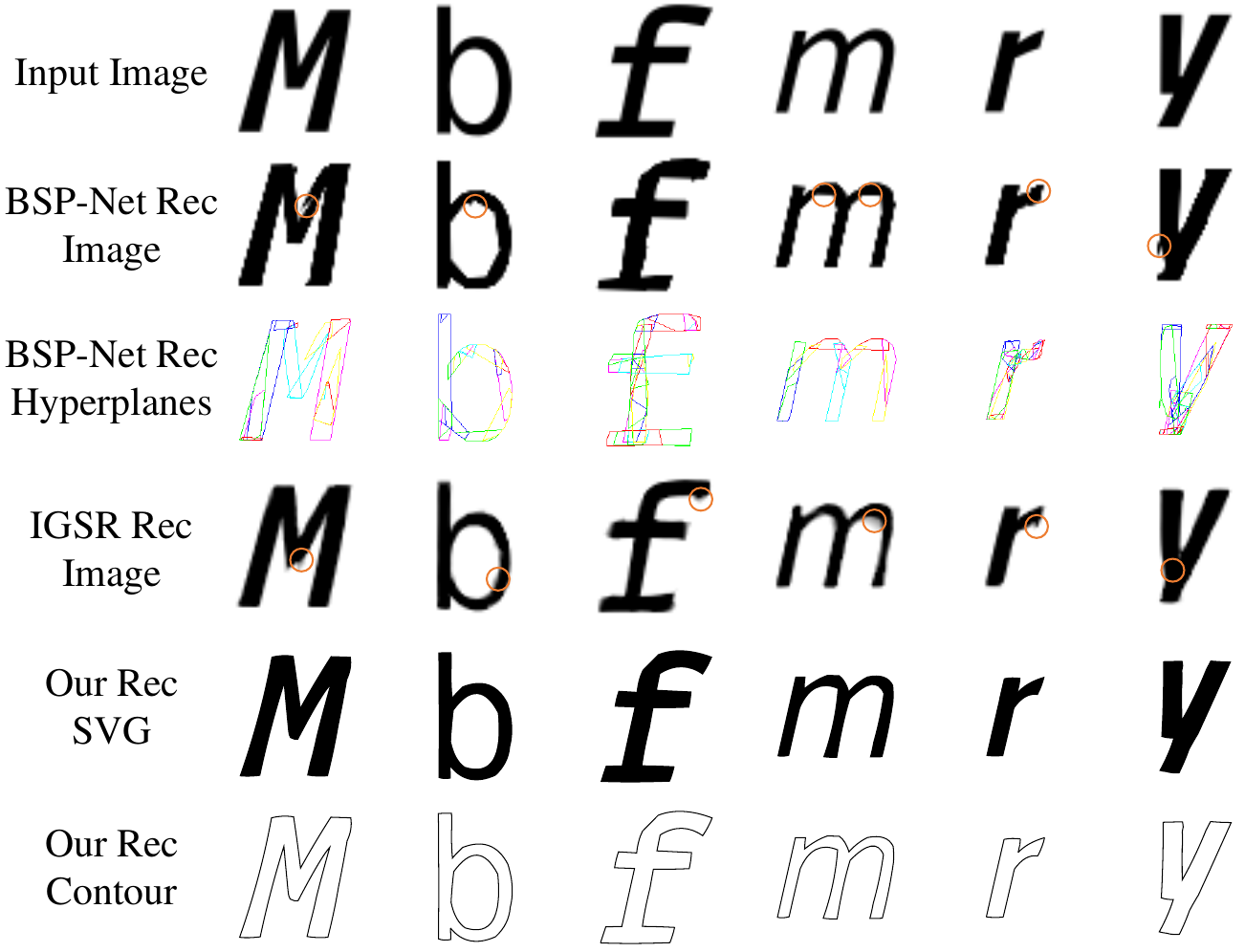}
  \caption{The qualitative comparison of our method and other SDF-based methods~\cite{chen2020bspnet, 9797843}. ``Rec" denotes ``Reconstructed". Orange circles highlight the poor performance of previous methods.}
  \label{fig:compare}
\end{figure}

\begin{figure}[t!]
  \centering
  \includegraphics[width=.9\columnwidth]{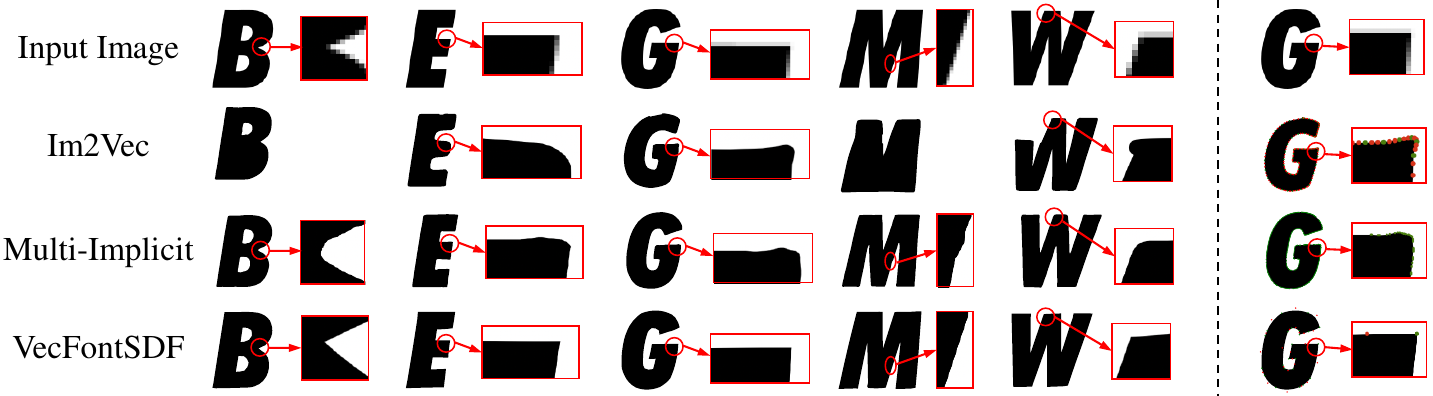}
  \caption{Vector reconstruction results of Im2Vec~\cite{reddy2021im2vec}, multi-implicits~\cite{NEURIPS2021_6948bd44} and our method. Please zoom in for more details.}
  \label{fig:compare-rebuttal}
\end{figure}

\subsection{Vector Font Reconstruction}

Fig.~\ref{fig:rec} shows the vector font reconstruction results of our method on the test set and the corresponding glyph contours after post processing, from which we can see that almost all vector glyphs reconstructed by our VecFontSDF are approximately identical to the corresponding input glyph images. Furthermore, our reconstructed vector fonts have the same visual effects as the human-designed vector fonts, without the loss of smoothness and shape deformation.

In Fig.~\ref{fig:compare}, we compare our method with two other existing SDF representations~\cite{chen2020bspnet, 9797843}. Since the original image resolution used in these two methods is 64 $\times$ 64, for fair comparison, we also resize the resolution of our input images and the size of sampling grid to $64 \times 64$. From Fig.~\ref{fig:compare} we can see that BSP-Net~\cite{chen2020bspnet} only uses straight hyperplanes which performs poorly on glyph images containing a large number of Bézier curves. IGSR~\cite{9797843} uses the general conic section equation $ax^2 + bxy + cy^2 + dx + ey + f=0$ to depict every curve which can not be precisely converted to the quadratic Bézier curve. What's more, both of them are only supervised by the $L_2$ image loss, leading to blur and imprecise reconstruction results.

\begin{table}[t!]
    \centering
    \caption{The comparison of quantitative results obtained by our method, BSP-Net~\cite{chen2020bspnet} and IGSR~\cite{9797843} on the reconstruction task.}
    
    \label{tab:comparison}
    \centering%
    \resizebox{\columnwidth}{!}{
    \begin{tabular}{lccccc}
      \toprule
      Method & $\rm L_1$ distance $\downarrow$ & IoU$\uparrow$ & PSNR$\uparrow$ & LPIPS$\downarrow$ &  SSIM$\uparrow$ \\
      \midrule
      BSP-Net~\cite{chen2020bspnet} & 0.0194 & 0.9854 & 22.4926 & 0.0500 & 0.9335 \\
      IGSR~\cite{9797843} & 0.0161 & 0.9837 & 24.6895 & 0.0304 & 0.9478 \\
      VecFontSDF (64 $\times$ 64) & \textbf{0.0120} & \textbf{0.9866} & \textbf{28.6739} & \textbf{0.0174} & \textbf{0.9662} \\
      \bottomrule
    \end{tabular}}
\end{table}

Table~\ref{tab:comparison} shows the quantitative results of different methods calculated using the $L_1$ distance, IoU, PSNR, LPIPS and SSIM, further demonstrating the effectiveness and superiority of our VecFontSDF to the state of the art.

We also compare our method with two recently-proposed vector reconstruction approaches: Im2Vec~\cite{reddy2021im2vec} and multi-implicits~\cite{NEURIPS2021_6948bd44} in Fig.~\ref{fig:compare-rebuttal}. Im2Vec~\cite{reddy2021im2vec} prefers to first fit the big outlines but sometimes ignores the small ones in glyphs (e.g., ``B" in column 1). It also tends to be stuck in local optimums for glyphs with multiple concave regions (e.g., ``M" in column 4). Multi-implicits~\cite{NEURIPS2021_6948bd44} results in finer details than Im2Vec but both of them exhibit unsmooth edges and rounded corners. What's more, Im2Vec produces hundreds of cubic Bézier curve control points to compose the glyph contour while most of them are redundant. Multi-implicits even needs to find thousands of points on the zero level-set of 2D SDF to vectorize the output. Thus, both of them are not suitable to generate practical vector fonts. On the contrary, glyphs synthesized by our method only contain dozens of quadratic Bézier curves which are more similar to human-designed ones. Detailed vector results containing all the control points are shown in the last column of Fig.~\ref{fig:compare-rebuttal}, where green points denote on-curve control points and red points denote off-curve control points.

\subsection{Vector Font Interpolation}
We also conduct experiments on vector glyph interpolation to further prove the ability of our representation. Given two input glyph images of the same character, we use the pre-trained image encoder on vector font reconstruction to extract the corresponding latent codes $z_1, z_2$. Then, an interpolated feature of them can be computed by $z = (1-\lambda)\cdot z_1 + \lambda \cdot z_2$.

\begin{figure}[t!]
  \centering
  \includegraphics[width=\columnwidth]{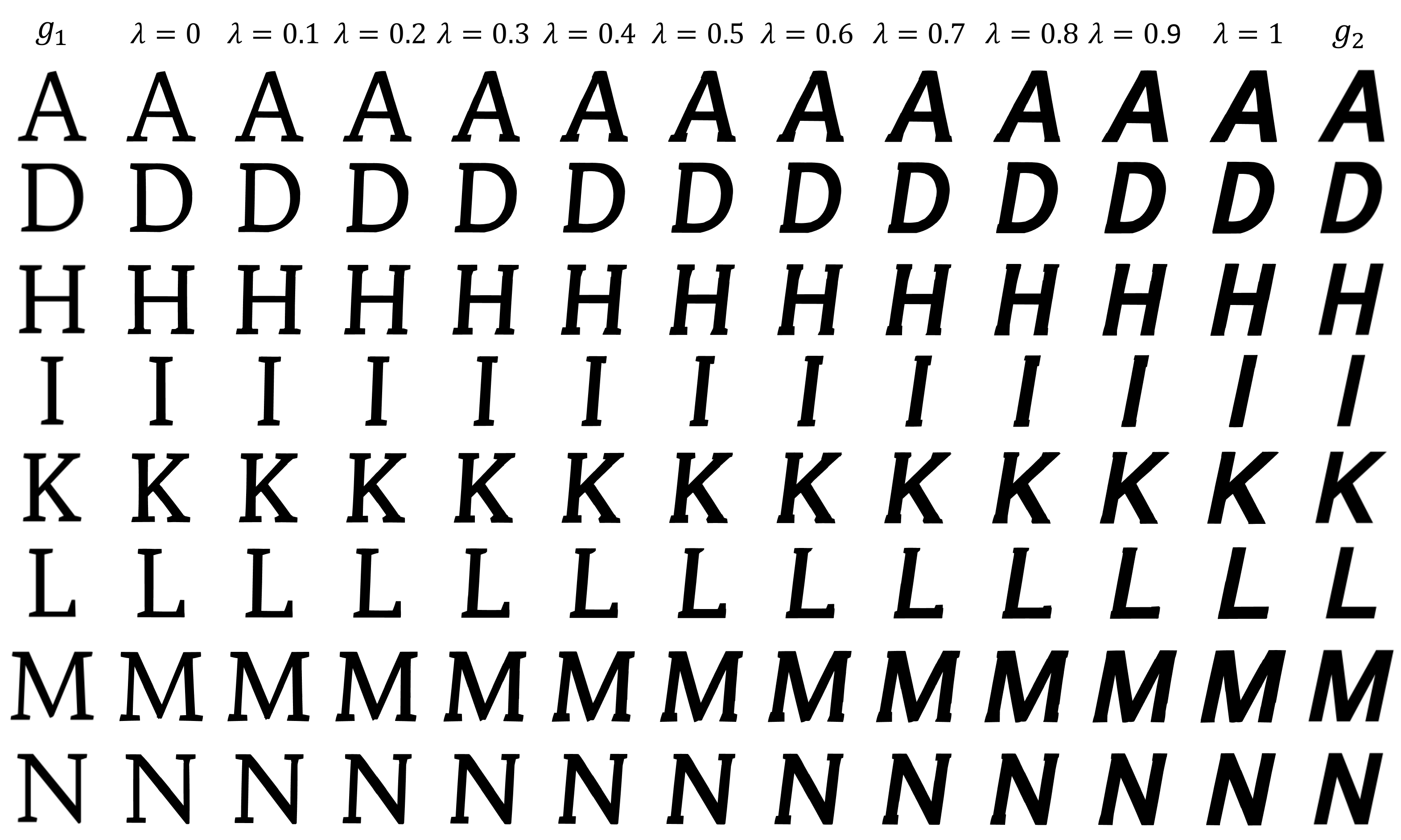}
  \caption{Examples of vector font interpolation results obtained by VecFontSDF. Our model can generate a series of high-quality vector fonts in new styles by only providing raster glyph images in two different font styles (the first and last columns).}  \label{fig:interpolation}
\end{figure}

\begin{figure}[t!]
  \centering
  \includegraphics[width=\columnwidth]{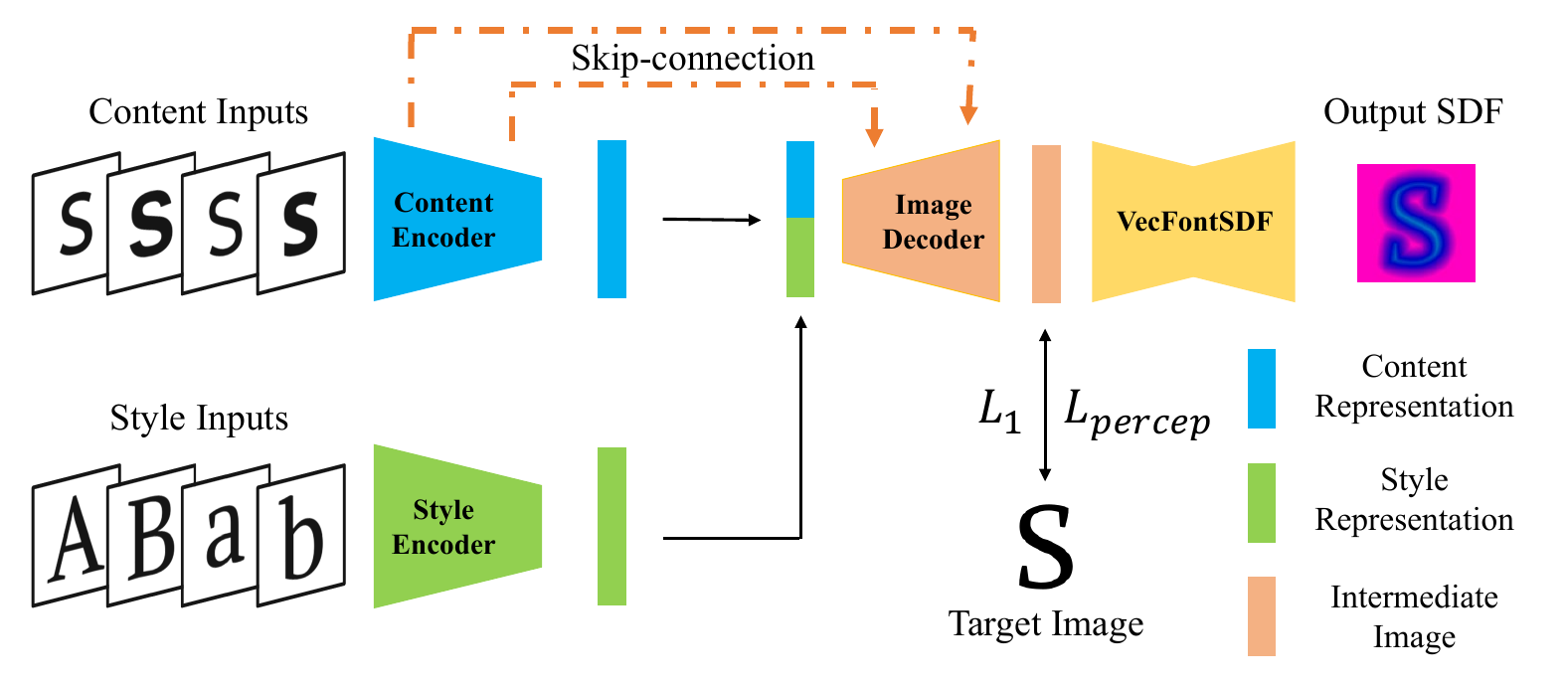}
  \caption{The pipeline of our few-shot style transfer model.}
  \label{fig:few-shot-network}
\end{figure}

We feed the latent code $z$ into our pre-trained SDF decoder and obtain the interpolated vector glyphs after post processing. Fig.~\ref{fig:interpolation} shows that our model achieves smooth interpolation between different styles of fonts and is capable of generating visually-pleasing new vector fonts by only inputting raster glyph images.

\begin{figure*}[t!]
  \centering
  \includegraphics[width=\textwidth]{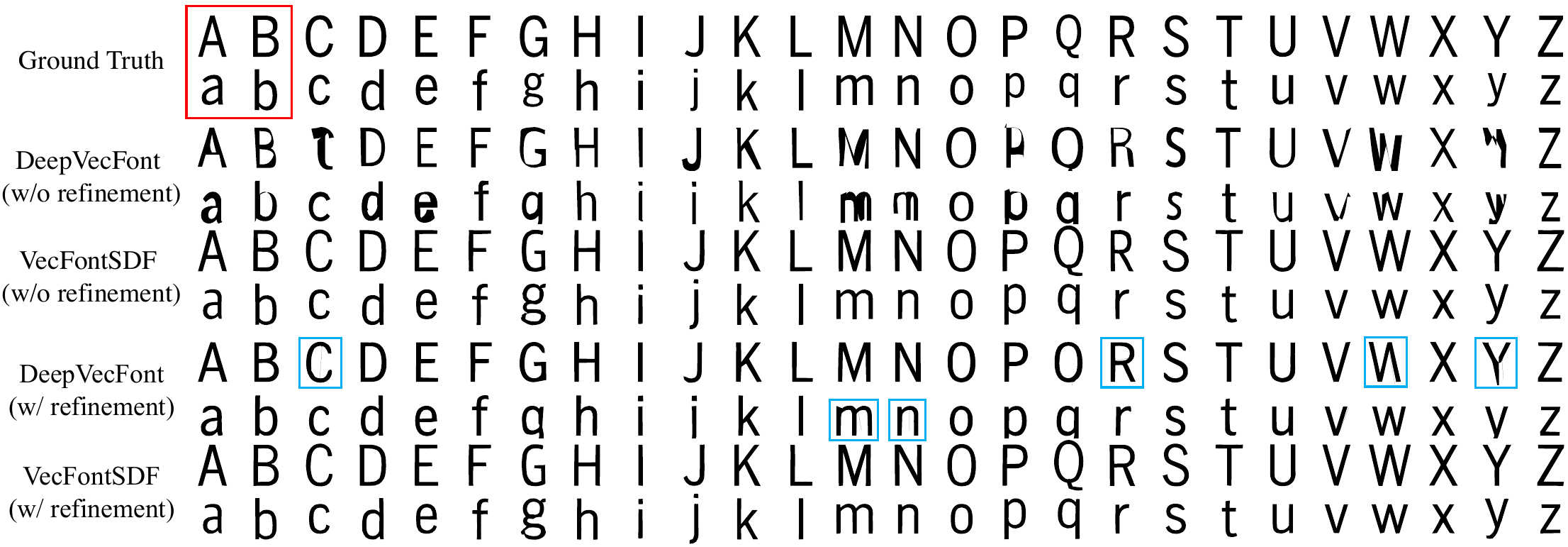}
  \caption{Few-shot vector font generation results of our VecFontSDF and a state-of-the-art method DeepVecFont~\cite{wang2021deepvecfont} before and after refinement. The input reference glyphs are marked by red rectangles. Note that our model only needs the four glyph images as input while DeepVecFont also needs these vector glyphs. Please zoom in for better inspection.}
  \label{fig:few-shot-compare}
\end{figure*}

\subsection{Few-Shot Style Transfer}
To further demonstrate the potential of our VecFontSDF for vector font generation, we directly extend the popular few-shot style transfer task from the image domain to the vector graphic domain. The input of our few-shot style transfer model consists of the style reference set $\mathcal{R}_{S_i}$ and the content reference set $\mathcal{R}_{C_i}$. Our goal is to output the vector glyph which has the same style as $\mathcal{R}_{S_i}$ and the same content as $\mathcal{R}_{C_i}$. The network architecture is shown in Fig.~\ref{fig:few-shot-network}.

\begin{figure}[t!]
  \centering
  \includegraphics[width=\columnwidth]{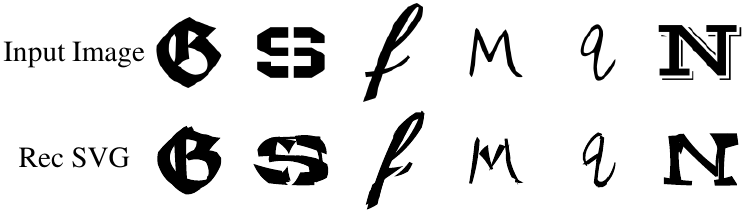}
  \caption{Typical failure cases of our method when handling glyphs with irregular shapes. ``Rec" denotes ``Reconstructed".}
  \label{fig:limitation}
\end{figure}

First, we use two separated CNN encoders to extract the style feature $z_s$ of $\mathcal{R}_{S_i}$ and the content feature $z_c$ of $\mathcal{R}_{C_i}$. Then we simply concatenate $z_s$ and $z_c$ and send it into the image decoder to get an intermediate image $\hat X_{c}$. We add skip connections between the content encoder and the image decoder like U-net~\cite{ronneberger2015u}. Our pretrained VecFontSDF obtained on the vector reconstruction task is followed to process the intermediate image and output corresponding SDFs. The intermediate image is supervised by $L_m$ which consists of the $L_1$ loss and the perceptual loss~\cite{johnson2016perceptual} compared with the groud-truth image $X_c$. The objective function of our model is the sum of $L_m$ and $L_{total}$ which is defined in Eq.~\ref{equ:whole-loss}. Our few-shot style transfer model is end-to-end trainable since the proposed VecFontSDF is differentiable. The two CNN encoders and image decoder use the residual block~\cite{he2016deep} with layer normalization~\cite{DBLP:journals/corr/BaKH16}, and the same optimizer configuration as mentioned in Sec.~\ref{sec:experiments} is adopted.

In Fig.~\ref{fig:few-shot-compare}, we compare the performance of our few-shot style transfer model with DeepVecFont~\cite{wang2021deepvecfont} on our test set. DeepVecFont receives both raster images and vector graphics as input while our model only needs raster images.  Moreover, our model is an end-to-end framework in the reference stage without offline refinement. Therefore, we compare our synthesized vector glyphs with the results of DeepVecFont before and after refinement. From Fig.~\ref{fig:few-shot-compare} we can see that the advantage of the explicit representation of DeepVecFont is that it outputs relatively smooth glyph contours. However, due to the difficulty of handling the long-range dependence, DeepVecFont faces serious shape distortion problems. On the contrary, our method simultaneously outputs all the parameters of parabolic curves which effectively avoids this issue. Although Diffvg~\cite{li2020differentiable} has a strong ability to align the generated vector glyphs with the target images, the refinement performance relies heavily on the quality of input vector glyphs. Due to the severe distortions and artifacts in some synthesized initial glyphs, there still exist many failure cases of DeepVecFont, where some thin lines and even artifacts may appear after refinement (glyphs marked in blue boxes and it would be better to open with Adobe Acrobat to see these thin lines). It can be concluded that our few-shot vector font synthesis method based on VecFontSDF markedly outperforms the state of the art, indicating the broad application of VecFontSDF.

\subsection{Failure Cases}
Although VecFontSDF enables high-quality vector font reconstruction under most circumstances, it still has difficulty to model complex glyph shapes occasionally. As shown in Fig.~\ref{fig:limitation}, our method synthesizes inaccurate vector glyphs for some strange shapes, like ``G" and ``S". It also faces degradation on some cursive glyphs like ``f", ``M" and ``q". What's more, it is also a tough task for VecFontSDF if input glyph images have a lot of disconnected regions like ``N" in the last column. This is mainly because our model struggles to cover such complicated geometries using a limited number of shape primitives.

\section{Conclusion}
In this paper, we presented a novel vector font shape representation, VecFontSDF, which models glyphs as shape primitives enclosed by a set of parabolic curves that can be translated to commonly-used quadratic Bézier curves. Experiments on vector font reconstruction and interpolation tasks verified that our VecFontSDF is capable of handling concave curves and synthesizing visually-pleasing vector fonts. Furthermore, experiments on few-shot style transfer demonstrated the ability of our VecFontSDF for many generation tasks. In the future, we are planning to upgrade our model by utilizing more effective network architectures to address the above-mentioned problems of our model when handling glyphs with irregular shapes, and extend our Pseudo Distance Functions to higher-order curves.

{\small
\bibliographystyle{ieee_fullname}
\bibliography{egbib}
}

\clearpage
\appendix

\section{Details of Post-processing Steps}

\subsection{From Parabolic Curves to a Shape Primitive}
The raw input of VecFontSDF consists of the parameters of parabolic curves. As mentioned in the main manuscript, each parabolic curve is defined by:
\begin{equation}
    k(px+qy)^2+dx+ey+f=0,
    \label{equ:parabolic_curve}
\end{equation}
and the inside area of the parabolic curve is defined as:
\begin{equation}
    H(x, y) = k(px+qy)^2+dx+ey+f < 0.
    \label{equ:inside_area}
\end{equation}
To reconstruct the geometry of an input glyph image, we first need to compute the intersection of $N_a$ areas to get a shape primitive. To realize this, we start from a initial square canvas from the left-bottom point (-1,-1) to the right-top point (1,1). The four sides of this initial square canvas can be treated as four special quadratic Bézier curves which degenerate into straight lines.

\begin{figure}[t!]
  \centering
  \includegraphics[width=\columnwidth]{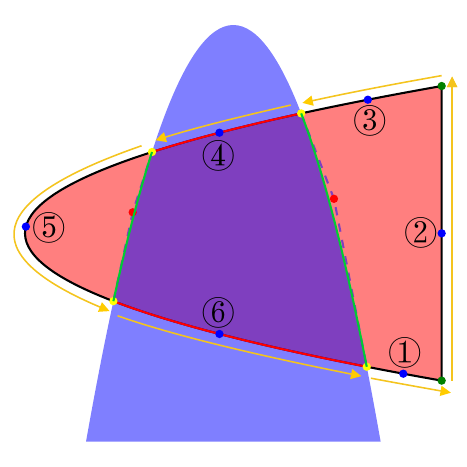}
  \caption{A detailed illustration of how to calculate the intersection of the current canvas and a newly-added parabolic curve.}
  \label{fig:primitive}
\end{figure}

As shown in Fig~\ref{fig:primitive}, when adding a new parabolic curve (one of the $N_a$ parabolic curves), we need to calculate the intersection of the inside area of the newly-added parabolic curve and current canvas. The current canvas (the red area) is enclosed by several quadratic Bézier curves and the inside area of newly-added parabolic curve (the blue area) is depicted by Eq.~\ref{equ:inside_area}. Therefore, the key problem to be resolved here is how to calculate the intersection points of a quadratic Bézier curve and a parabolic curve. Recall that, a standard representation for the quadratic Bézier curve can be described as:
\begin{equation}
\begin{split}
    {P(t)}&=(1-t)^2P_0+2t(1-t)P_1+t^2P_2 \\ 
    0&\leq t\leq 1.
\end{split}
\label{equ:basic_bezier}
\end{equation}
We simply substitute the $x$ and $y$ in Eq.~\ref{equ:parabolic_curve} with the coordinates of the points in Eq.~\ref{equ:basic_bezier} to get a new quartic equation of $t$. After solving this quartic equation and validating $0 \leq t \leq 1$, we can obtain all the intersection points (the yellow points in Fig~\ref{fig:primitive}).

Then, we need to update the current canvas (the red area) to the newly-calculated intersection region (the purple area). The newly-calculated intersection points (four yellow points) divide the current canvas into six segments as marked in Fig.~\ref{fig:primitive}. Due to the divisibility of Bézier curves, a given Bézier curve
\begin{equation}
    B:P(t)=\sum_{i=0}^{n}{\left[ C_n^it^i(1-t)^{n-i}P_i \right] }
    \label{equ:any_bezier}
\end{equation}
can be divided by any point $\hat t$ on the curve into two parts and every part is still a Bézier curve with the same order as the original one. For example, the left part $B_l$ can be defined as:
\begin{equation}
    \begin{split}
    B_l:P_l(t)&=\sum_{i=0}^{n}{\left[C_n^it^i(1-t)^{n-i}\hat{P_i}\right]} \\ \hat{P}_i&=\sum_{j=0}^{i}{\left[C_i^j\hat{t}^j(1-\hat{t})^{i-j}P_j\right]},
    \end{split}
\end{equation}
and so is the right part. Therefore, these six segments are also quadratic Bézier curves. 

To accurately obtain this intersection region (the purple area), we start from one of these intersection points (four yellow points), for example, the right-bottom yellow point. Then, we traverse anticlockwise through all the segments one by one (as pointed out by the yellow arrows in Fig.~\ref{fig:primitive}) and judge whether every segment is inside the parabolic curve (the blue area). Due to the convexity of quadratic Bézier curves, we only need to calculate if its middle point ($t=0.5$) (the blue point) is inside this area. For the case shown in Fig.~\ref{fig:primitive}, the first three segments are not inside the blue area, so we simply ignore them and go to the right-top yellow point. The fourth segment (the red segment on the top) is exactly inside the blue area, so we reserve this segment and calculate the quadratic Bézier curve (a part of the parabolic curve that is marked as the green segment on the right) whose two on-curve control points are the current yellow point (right-top) and the previous yellow point (right-bottom).

\begin{figure}[t!]
  \centering
  \includegraphics[width=\columnwidth]{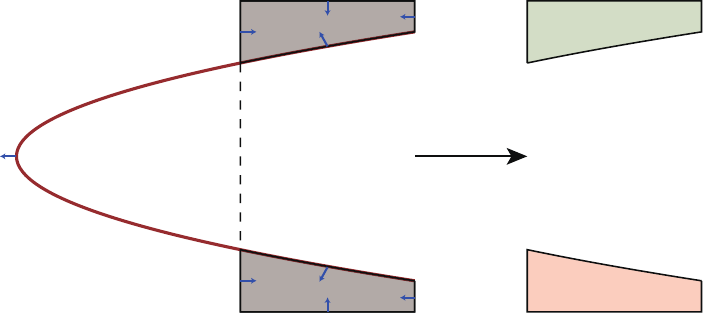}
  \caption{A typical corner case: a newly-added parabolic curve (the red curve on the left) splits the existing canvas into multiple parts (right). Blue arrows denote the directions of the inside areas from Eq. \ref{equ:inside_area}.}
  \label{fig:slim}
\end{figure}

Since the intersection point of the tangents of two on-curve control points of a quadratic Bézier curve is exactly the off-curve control point, we can calculate the equations of these two tangents (the purple dashed lines on the right in Fig.~\ref{fig:primitive}) to get the coordinate of the off-curve control point (the red point on the right). Based on Eq.~\ref{equ:parabolic_curve}, the slope $y'$ of tangent on any point $(x_0, y_0)$ can be calculated as:
\begin{equation}
    2k\left(px_0+qy_0\right)\left(p+qy'\right)+d+ey'=0,
\end{equation}
which equals to
\begin{equation}
    (2k(px_0+qy_0)q+e)y'=-(2k(px_0+qy_0)p+d).
    \label{equ:tangent_slope}
\end{equation}
Therefore, the equation of the tangent on $(x_0, y_0)$ is:
\begin{equation}
    y-y_0=y'(x-x_0),
    \label{equ:tangent_point}
\end{equation}
where $y'$ can be solved from Eq.~\ref{equ:tangent_slope}.

After getting the two equations of tangents on two on-curve control points, we can use them to calculate the coordinate of the off-curve control point (the red point on the right). As shown in Fig.~\ref{fig:primitive}, we repeat the above process on the fifth and sixth segments where the sixth segment is inside the blue area to get another quadratic Bézier curve (the green segment on the left). Finally, we use these four quadratic Bézier curves (two red segments and two newly-calculated green segments) to update the current canvas.

\begin{figure}[t!]
  \centering
  \includegraphics[width=\columnwidth]{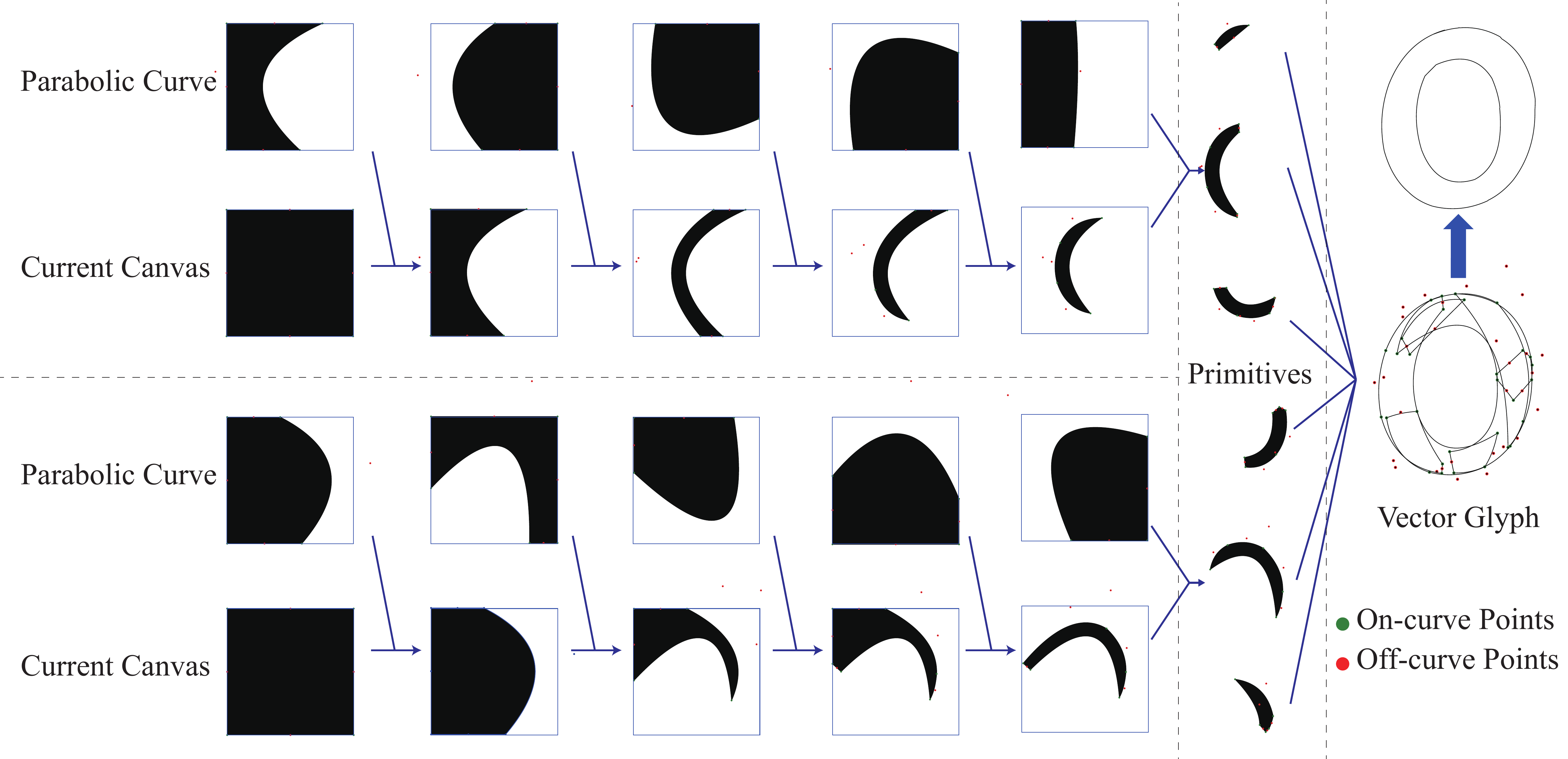}
  \caption{The pipeline of our post processing step (also shown as Fig. 4 in the main manuscript).}
  \label{fig:postproc}
\end{figure}

We implement the above-mentioned process recursively, and eventually get the primitive as shown in Fig.~\ref{fig:postproc}.

It is notable that Fig.~\ref{fig:slim} shows a typical corner case, where a newly-added parabolic curve (the red curve in Fig.~\ref{fig:slim}) splits the existing canvas (the gray area on the left) into several parts (the green and red areas on the right). Under this circumstance, we need to check whether the middle point of every segment from newly-added parabolic curve is out of the existing canvas. If so, we separate the existing canvas into the corresponding parts, and handle each part individually.

\subsection{Calculating Intersection Points of Two Quadratic Bézier Curves}
\label{sec:ib_qbc}
As shown in Fig.~\ref{fig:postproc}, after obtaining all the primitives, we can get the filled shape of the target vector glyph by simply assembling them together with many intersections inside the contour. However, further simplifying the primitives' outlines is necessary to obtain a complete outline representation of the vector glyph. Before we introduce the details of how to merge all the outlines of primitives, we need to clarify the process of calculating the intersection points of two quadratic Bézier curves which plays an important role in the following section.

To calculate all the possible intersection points of two quadratic Bézier curves described by Eq. \ref{equ:basic_bezier}, we have
\begin{equation}
\begin{split}
    {\exists} \quad t_1,t_2&\in[0,1] \\
    {\rm s.t.} \quad x_1\left(t_1\right)&=x_2\left(t_2\right) \\
    y_1\left(t_1\right)&=y_2\left(t_2\right).
\end{split}
\label{equ:bezier_cross_init}
\end{equation}

The left hand side of Eq. \ref{equ:bezier_cross_init} equals to:
\begin{equation}
\begin{split}
    (1-t_1)^2x_{1,0}+2t_1(1-t_1)x_{1,1}+t_1^2x_{1,2}&=x_2\left(t_2\right) \\
    (1-t_1)^2y_{1,0}+2t_1(1-t_1)y_{1,1}+t_1^2y_{1,2}&=y_2\left(t_2\right).
\end{split}
\label{equ:bezier_cross_expand_1}
\end{equation}

We expand the above equation and let:
\begin{equation}
\begin{split}
    A&=x_{1,0}-2x_{1,1}+x_{1,2} \\
    B&=-2x_{1,0}+2x_{1,1} \\
    C&=x_{1,0} \\
    D&=y_{1,0}-2y_{1,1}+y_{1_2} \\
    E&=-2y_{1,0}+2y_{1,1} \\
    F&=y_{1,0}.
\end{split}
\label{equ:bezier_cross_expand_parameters}
\end{equation}

Thus, we can rewrite the Eq.~\ref{equ:bezier_cross_expand_1} into:
\begin{align}
    At_1^2+Bt_1+C&=x_2\left(t_2\right) \label{equ:bezier_cross_expand_x}\\
    Dt_1^2+Et_1+F&=y_2\left(t_2\right)\label{equ:bezier_cross_expand_y}.
\end{align}

Let (\ref{equ:bezier_cross_expand_x})$\times D -$ (\ref{equ:bezier_cross_expand_y}) $\times A$, we get
\begin{equation}
\begin{split}
    &(DB-AE)t_1+(DC-AF) \\
    =&Dx_2\left(t_2\right)-Ay_2\left(t_2\right).
\end{split}
\label{equ:bezier_crosses_diffed}
\end{equation}

The left hand side of Eq.\ref{equ:bezier_crosses_diffed} is a linear expression of $t_1$, and the right hand side is a quadratic expression of $t_2$.

Assume $DB-AE \neq 0$, we can represent $t_1$ by a quadratic function of $t_2$:
\begin{equation}
    t_1=\frac{Dx_2\left(t_2\right)-Ay_2\left(t_2\right)-DC+AF}{D
    B-AE}.
    \label{equ:t1_express}
\end{equation}

Then, we substitute $t_1$ in Eq.~\ref{equ:bezier_cross_expand_1} with Eq. \ref{equ:t1_express} and thus we can get a quartic equation of $t_2$. Afterwards, we solve this equation to get $t_2$ and use Eq.~\ref{equ:t1_express} to get $t_1$.

Otherwise, if $DB-AE=0$, we directly solve the quadratic equation of $t_2$ from Eq.~\ref{equ:bezier_crosses_diffed} and use this already known $t_2$ to solve one of the equations in Eq. \ref{equ:bezier_cross_expand_1} to get $t_1$.

After calculating $t_1$ and $t_2$ and validating if both of them satisfy the constraint $0 \leq t \leq 1$, we can finally use Eq. \ref{equ:basic_bezier} to calculate the intersection points of these two quadratic Bézier curves.

\begin{figure}[t!]
  \centering
  \includegraphics[width=\columnwidth]{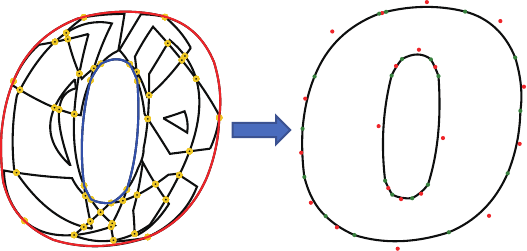}
  \caption{An illustration of our outline simplification step. Yellow points denote the intersection points of all quadratic Bézier curves. The red outline and blue outline compose this glyph's contour. Black segments are all other quadratic Bézier curves whose middle points' SDF values are less than 0 (i.e., inside the glyph contour).}
  \label{fig:simple_o}
\end{figure}

\subsection{Outline Simplification} 
\label{sec:outline_simple}

As shown in Fig.~\ref{fig:simple_o}, to merge the outlines of all primitives, we first need to calculate the intersection points (yellow points in Fig.~\ref{fig:simple_o}) of all existing quadratic Bézier curves using the method described in Sec.~\ref{sec:ib_qbc}. After calculating all the intersection points, every quadratic Bézier curve is further divided into shorter segments. Due to the  divisibility of Bézier curves, these shorter segments are also quadratic Bézier curves.

For all the existing quadratic Bézier curves after division, we can directly ignore the ones inside the glyph contour (e.g., the black segments in Fig.~\ref{fig:simple_o}) and reserve the ones composing the glyph contour (all the quadratic Bézier curves lying on the red and blue contours). Recall that in the main manuscript we have already get the pseudo distance functions $G(x,y)$ to calculate the value of the pseudo distance on each sampling point $(x,y)$, which is an effective tool to help us determine the position of every segment. Due to the convexity of quadratic Bézier curves, whether a quadratic Bézier curve is contained in the glyph contour or not can be judged by the signed distance of its middle point ($t=0.5$) $G(x_m,y_m)$, where $(x_m,y_m)$ denotes the coordinates of the middle point.

For a given quadratic Bézier curve, if $G(x_m,y_m)=0$ which means it composes the glyph contour, we reverse this curve and discard the curves whose $G(x_m, y_m) < 0$ (the black segments). Finally, we obtain all the quadratic Bézier curves lying on the glyph contour (the red and blue contours) to finish our outline simplification step.

\begin{figure*}[t!]
  \centering
  \includegraphics[width=\textwidth]{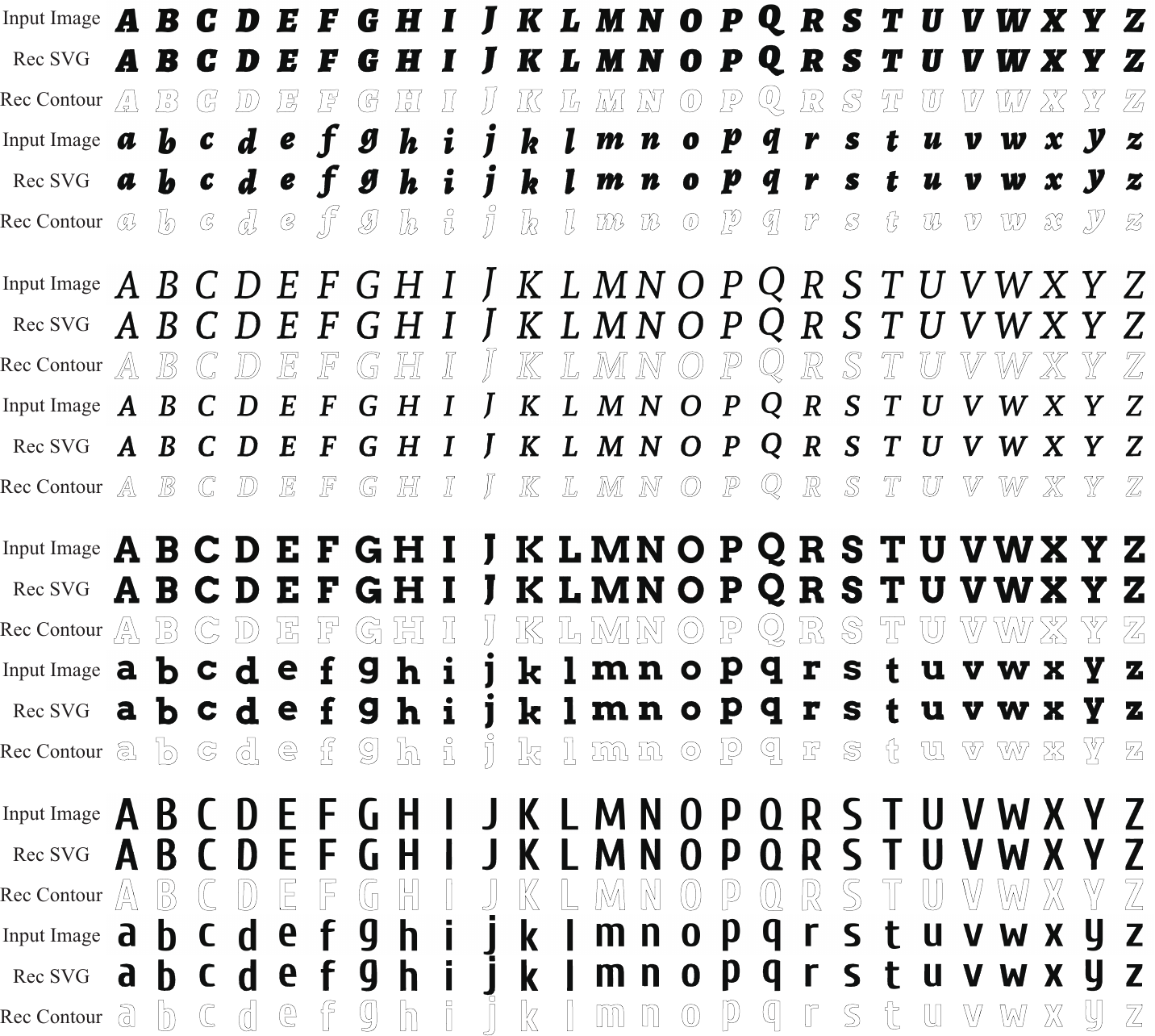}
  \caption{More vector font reconstruction results obtained by our method. ``Rec SVG" denotes the filled shape of the reconstructed vector glyphs and ``Rec Contour" denotes the contours of corresponding vector glyphs.}
  \label{fig:rec_suppl}
\end{figure*}

\begin{figure*}[t!]
  \centering
  \includegraphics[width=\textwidth]{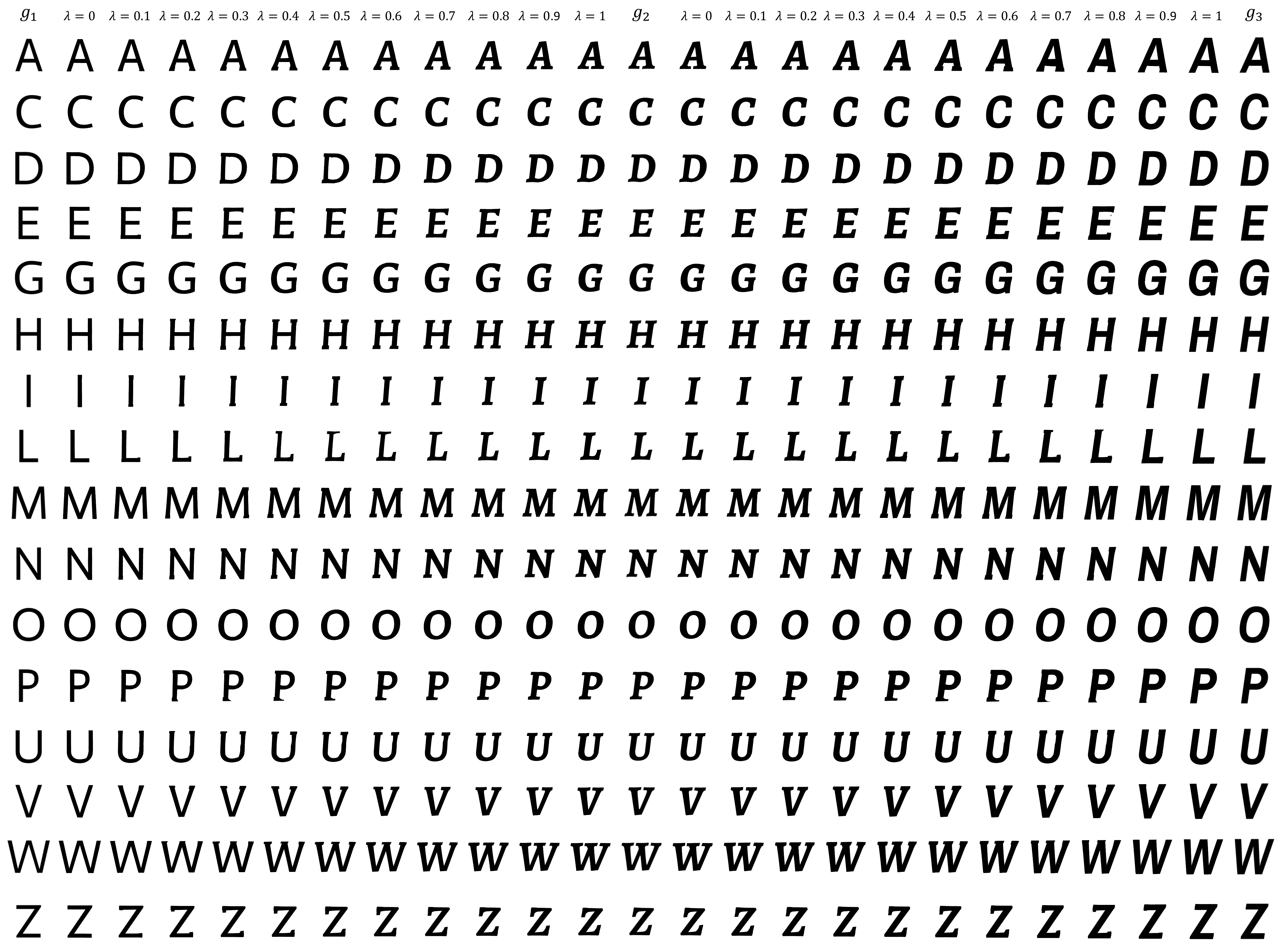}
  \caption{Our method is capable of generating vector fonts in new styles by only providing raster glyph images (the columns `$g_1$', `$g_2$' and `$g_3$') in different font styles. Two interpolation processes (`$g_1$' to `$g_2$' and `$g_2$' to `$g_3$') are presented in succession.}

  \label{fig:interpolation}
\end{figure*}


\section{More Results}
In this section, we provide additional experimental results in
support of the conclusions drawn in the main manuscript.

\subsection{Vector Font Reconstruction}
Fig.~\ref{fig:rec_suppl} shows more vector font reconstruction results obtained by our method as well as the corresponding glyph contours after implementing the simplification step mentioned in Sec.~\ref{sec:outline_simple}, from which we can see that our method is capable of reconstructing various styles of fonts given the input raster images, including some complex glyphs with serifs.

\subsection{Vector Font Interpolation}
Fig.~\ref{fig:interpolation} shows more results on vector font interpolation by only inputting raster glyph images (denoted as the columns `$g_1$', `$g_2$' and `$g_3$' in Fig.~\ref{fig:interpolation}). Our method achieves smooth interpolation between different styles of fonts demonstrating that the latent space embedded by our CNN encoder is perceptually smooth and interpretable to represent the vector font style.

\subsection{Few-shot Style Transfer}
\begin{figure*}[t!]
  \centering
  \includegraphics[width=\textwidth]{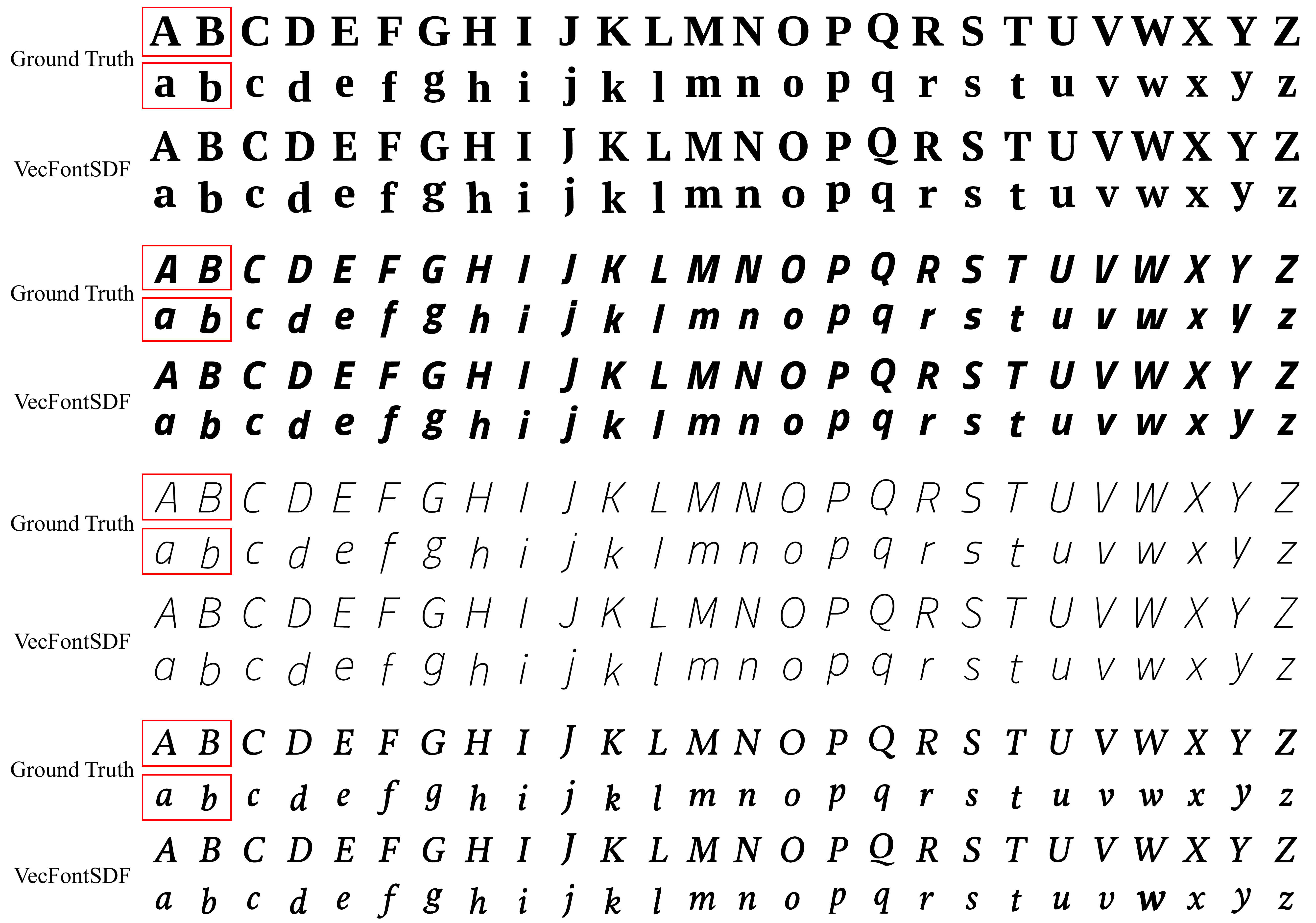}
  \caption{More few-shot vector font generation results of our VecFontSDF. The inputs to our networks are only the rendering results (raster images) of corresponding vector glyphs marked by red rectangles.} 

  \label{fig:few-shot}
\end{figure*}

Fig.~\ref{fig:few-shot} shows more few-shot style transfer results generated by our method. As shown in Fig.~\ref{fig:few-shot}, we demonstrate the effectiveness of our method in the task of generating all other vector glyphs by only giving a few reference glyph images instead of vector glyphs. The input reference glyphs are marked by red rectangles in Fig.~\ref{fig:few-shot}, from which we can see that our model is capable of synthesizing the whole vector font given just a small number of reference glyph images. Especially for the second and third fonts, almost all glyphs in the synthesized vector font are approximately identical to the human-designed vector glyphs. For some complex and complicated fonts with serifs (such as the first and fourth fonts in Fig.~\ref{fig:few-shot}), we observe that some local details of our synthesis results are only slightly different against the ground-truth glyphs. Considering that our model only receives the glyph images of ‘A’, ‘B’, ‘a’ and ‘b’ as reference inputs, our synthesized glyphs of other characters have already sufficiently embodied the style feature of these input samples. Most importantly, all the vector glyphs generated by our method are of high quality and look visually pleasing, which markedly outperform other existing state-of-the-art methods without further refinement.

\end{document}